\documentclass[letterpaper, 10 pt, conference]{ieeeconf}  

\IEEEoverridecommandlockouts                              

\usepackage{amsmath} 
\usepackage{amssymb}  
\usepackage{graphicx}
\usepackage{xcolor}

\title{\LARGE \bf
History-Aware Force Sensing and Tunable Stiffness in a Pneumatic Kresling Origami Cell
}

\author{Hengyi Zhou$^{1}$, Jiaqi Wang$^{1}$, Yilin Ma$^{1}$,
Dezhong Tong$^{1}$, Xiaonan Huang$^{1,*}$%
\thanks{$^{1}$Authors are with the University of Michigan, Ann Arbor, USA.}%
\thanks{$^{*}$Corresponding author: Xiaonan (Sean) Huang.
        Email: {\tt\small xiaonanh@umich.edu}.}%
}

\usepackage[hidelinks]{hyperref}
\hypersetup{
    pdfauthor={},
    pdfsubject={},
    pdfkeywords={}
}

\begin{document}

\maketitle
\thispagestyle{empty}
\pagestyle{empty}

\begin{abstract}
Mechanical hysteresis and pneumatic loading complicate contact force estimation from deformation measurements in soft robots. This work presents a 3D-printed polyether block amide (PEBA) Kresling cell with magnetic angular sensing that supports three operating modes: force sensing at zero gauge pressure, force sensing with tunable stiffness under gauge positive pressure, and self-sensing actuation under negative gauge pressure. A multiscale gated recurrent unit (GRU) network estimates contact force from angular histories to account for the the combined effects of geometric deformation and material response. At zero gauge pressure, the model achieves a test RMSE of 0.76~N. Positive gauge pressure approximately doubles the average axial loading stiffness at +20~kPa over the first 3~mm of compression, while pressure-specific models maintain RMSEs below 0.93~N. Under negative gauge pressure, an energy-based compensation model enables contact force estimation using the the model trained at zero gauge pressure, achieving RMSEs of 1.66--2.10~N under quasi-static conditions. These results demonstrate a compact approach to combining force feedback, tunable axial stiffness, and self-sensing actuation in soft origami systems.

\end{abstract}

\section{Introduction}

Proprioception and force feedback enable soft robots to monitor deformation and regulate physical interactions. Their large deformations require compatible sensing approaches, including stretchable optical waveguides, embedded conductive sensors, and directly printed resistive sensors \cite{zhao2016optoelectronic,truby2018somatosensitive,shih2019sensors}.

Origami-inspired structures have been explored as both soft actuators and sensors. Their folding patterns provide geometric mechanisms for converting pneumatic inputs into contraction, bending, and twisting motions \cite{jin2022origami}. In particular, Kresling origami couples axial contraction with rotation, supporting integrated actuation and sensing \cite{masana2024kresling}. Capacitive sensing \cite{hanson2024fold,oliveira2026orisense}, rotation measurements \cite{fan2024vacuum,fan2025sparc}, and conductive organogels \cite{wei2025biodegradable} have enabled deformation monitoring and proprioceptive feedback. Related studies have demonstrated optical and triboelectric force measurement \cite{wang2023coiling,kim2026bifurcation}, and continuous force estimation in modular origami robots and soft pneumatic modules \cite{huang2022modular,buso2020contact}.

However, force sensing in pneumatic Kresling cells presents two challenges. First, geometric and material effects introduce history dependence into the mechanical response \cite{mora2025landscape}. The Kresling cell exhibits pronounced hysteresis, with different loading and unloading forces at the same angular displacement, and a first-cycle effect, with higher loading forces in the first cycle of each test sequence than in subsequent cycles (Fig.~\ref{fig:0pressure}a). Consequently, instantaneous angular displacement alone is insufficient to determine contact force. Unmodeled loading history has also been identified as a limitation of force prediction in origami robots \cite{huang2022modular}. Second, pneumatic actuation contributes to the measured deformation, requiring its force contribution to be accounted for when estimating external contact force.

For history-dependent sensing, recurrent networks have been used in other soft systems to estimate contact pressure from hysteretic microfluidic sensor signals \cite{han2018microfluidic}, external forces on soft continuum actuators \cite{thuruthel2019perception}, and soft robot configuration from hysteretic piezoresistive signals \cite{truby2020distributed}. 

Inspired by these studies, we investigate history-aware force estimation from angular trajectories of a 3D-printed PEBA Kresling cell equipped with magnetic angular sensing. The same assembly is evaluated in three operating modes: force sensing at zero gauge pressure, force sensing with tunable stiffness under positive gauge pressure, and self-sensing actuation under negative gauge pressure. We explore positive pressure as a means of adjusting axial stiffness and the accessible force range within a prescribed compression interval, and investigate external contact force estimation during negative-pressure-driven contraction.

To account for history dependence, a multiscale GRU network combines the current angle with angular histories sampled at different frequencies \cite{cho2014gru}. These inputs provide both recent deformation changes and longer loading histories, allowing the network to implicitly learn the coupled effects of geometry and material properties on the force response. Separate models are trained at each tested positive pressure to account for the corresponding mechanical response. For negative-pressure actuation, an energy-based compensation model corrects the output of the network trained at zero gauge pressure (the baseline model) for the pressure-equivalent force, providing approximate contact force estimates under quasi-static conditions without retraining.

The main contributions of this work are:
\begin{itemize}
\item A multiscale, history-aware force estimator that captures nonlinear and history-dependent force responses from Kresling angular trajectories.
\item An experimental investigation of positive pressure tuning of axial stiffness and accessible force range, with force sensing enabled by pressure-specific models.
\item An energy-based pressure compensation model that enables quasi-static contact force estimation under negative gauge pressure using the baseline model without retraining.
\end{itemize}

\section{Mechanical Design \& Fabrication}
The rendered Kresling cell can be found in Fig.~\ref{fig:geo}a and the geometric design is shown in Fig.~\ref{fig:geo}b. The cell is defined by two regular hexagons with a circumradius of $r=15$~mm, an initial axial separation of $h_0=30$~mm, and an initial relative rotation of $\varphi=50^\circ$. Each of the six repeating lateral units comprises two triangular facets, $F_1$ and $F_2$, sharing a diagonal crease. The corresponding edge lengths are $a=15.00$~mm, $b=32.57$~mm, and $d=38.78$~mm. The planar angles of the unfolded triangular facets are $\alpha=54.95^\circ$, $\beta=102.90^\circ$, and $\gamma=22.15^\circ$, following the definitions in Fig.~\ref{fig:geo}b.

Flexible thermoplastics, including TPU \cite{hanson2024fold} and other TPE formulations \cite{jin2022origami,fan2025sparc}, have been used to fabricate Kresling actuators. Here, PEBA 90A was selected for its flexibility and high rebound resilience. Su et al. reported a rebound resilience of 66.5\% for PEBA 90A, compared with 32.0\% for the TPU tested in their study, and demonstrated its use in elastic metamaterials and soft actuators \cite{su2026peba}. These findings motivated its selection for a Kresling cell undergoing repeated compression and recovery. Its compatibility with fused filament fabrication also enables direct printing of the hollow cell geometry. 

Despite the high rebound resilience of PEBA, the printed Kresling cell still exhibits pronounced hysteresis and a first-cycle effect (Fig.~\ref{fig:0pressure}a). These observations indicate that deformation history must still be considered when estimating contact force from angular measurements.

The Kresling cell was 3D-printed from PEBA 90A using a Bambu Lab H2D printer. The wall thickness was 0.52~mm, and the top and bottom end caps were each 3.5~mm thick. The end caps provided interfaces for attachment to the mounting fixtures.

\section{Experimental Setup}
As-printed Kresling specimens may contain residual stresses and exhibit the Mullins effect \cite{mullins1969}. The specimens were subjected to 1,000 10~mm loading–unloading cycles for mechanical preconditioning.

\begin{figure}[t]
    \centering
    \includegraphics[width=\columnwidth]{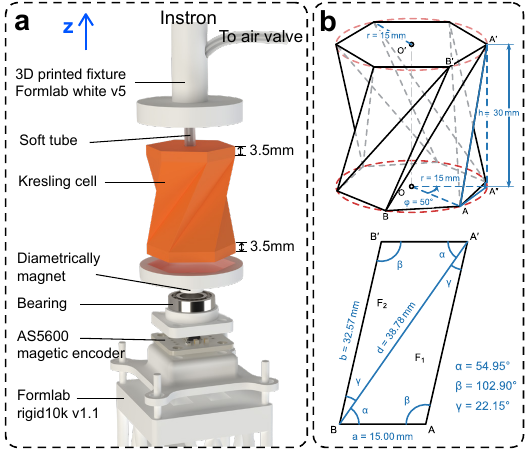}
    \caption{Experimental setup and Kresling geometry. (a) Experimental assembly with magnetic angular sensing. (b) Geometric parameters and unfolded triangular facets.}
    \label{fig:geo}
\end{figure}

In all three modes, the Instron force measurements were used as the ground truth, as shown in Fig.~\ref{fig:geo}a. To improve measurement accuracy, the bottom fixture was 3D-printed using Formlabs Rigid 10K V1.1 resin. White V5 resin was used for the remaining assembly fixtures to achieve a tight fit among the Kresling cell, fixtures, and bearing.

The AS5600 magnetic encoder and the bearing housing were constrained against rotation about the z-axis while remaining free to translate along it with negligible friction. A diametrically magnetized magnet is fixed in the bottom cover, rotating with the Kresling cell. A soft tube passes through the top fixture and connects the Kresling chamber to a positive or negative pneumatic valve, which is connected to either a compressed air source or a vacuum pump.

As the Instron moves downward to compress the Kresling cell, the bottom of the cell is free to rotate about the z-axis, and the angle is recorded continuously by the magnetic encoder. The magnetic encoder measures the absolute angular position of the diametrically magnetized magnet, which is converted into an angular displacement relative to the initial reference angle. All angles reported hereafter refer to this relative angular displacement.

To establish an initial reference point corresponding to approximately zero contact force, the Instron was programmed to compress the Kresling cell until the measured force reached 10~N and then unload at 0.1~mm/s. The point at which the measured force fell below 0.1~N was defined as the reference point. The maximum compression displacement was limited to 10 mm, and Abaqus simulations confirmed that no self-contact occurred within this range. 

Invalid regions are shaded gray in the following figures and excluded from analysis because they either exceed the safe compression range, where self-contact may occur, or contain frozen AS5600 encoder readings. Details are included in Finite Element Analysis part. 

\section{Method}

\subsection{Zero Gauge Pressure: GRU Network Force Sensing}

\begin{figure*}[t]
    \centering
    \includegraphics[width=\textwidth]{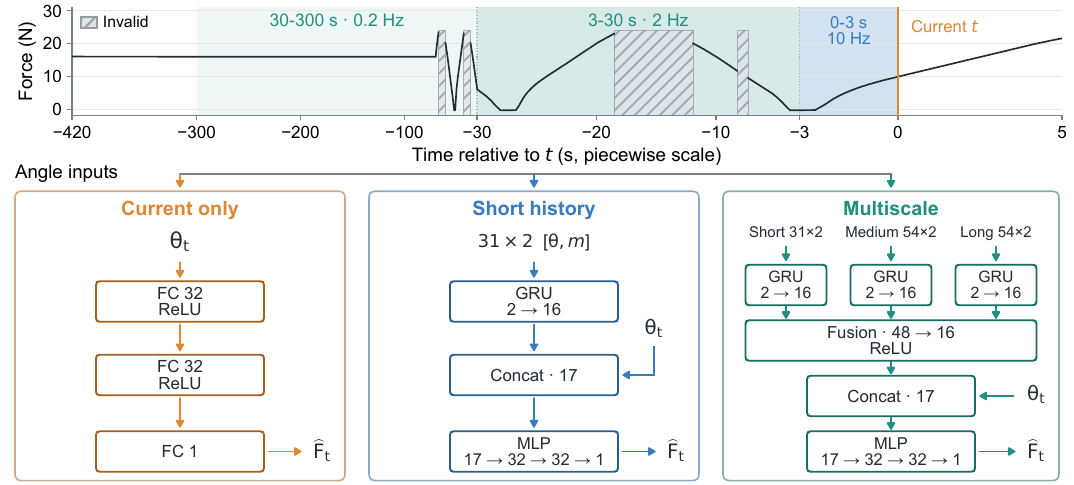}
    \caption{Force estimation architectures and temporal input windows. The current angle MLP uses only the instantaneous angle, whereas the short history and multiscale GRU networks additionally incorporate angular histories with validity indicator. The multiscale model combines short (0–3 s, 10 Hz), medium (3–30 s, 2 Hz), and long (30–300 s, 0.2 Hz) histories. The upper panel illustrates these windows relative to the current time (t) on a force trajectory with a piecewise time scale; hatched regions indicate invalid data.}
    \label{fig:archi}
\end{figure*}

\begin{figure*}[t]
    \centering
    \includegraphics[width=\textwidth]{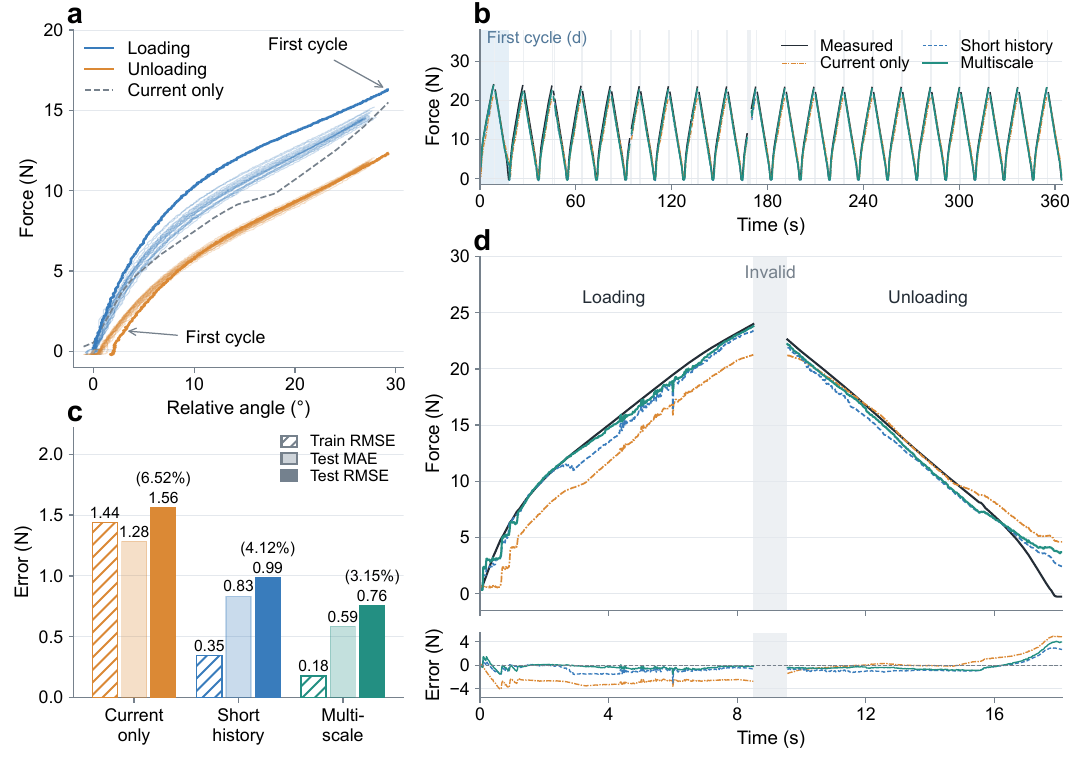}
    \caption{Force estimation performance at zero gauge pressure. (a) Force–angle responses showing hysteresis and first-cycle effects, with the current angle MLP prediction overlaid. (b) Measured and estimated force trajectories for the three models. (c) Comparison of training RMSE, test MAE, and test RMSE. (d) Detailed view of the first cycle highlighted in (b), with corresponding estimation errors. Gray shaded regions indicate invalid data excluded from analysis.}
    \label{fig:0pressure}
\end{figure*}

The geometry and material response of the Kresling origami structure introduce history dependence into its sensing behavior. The solid lines in Fig.~\ref{fig:0pressure}a show first 20 cycles in a test and the loading and unloading of the first cycle are thickened. First, hysteresis is observed, with distinct loading and unloading trajectories. At the same measured angular displacement, the force during unloading is consistently lower than that during loading. Second, a pronounced first-cycle effect persists even after 1,000 preconditioning cycles.

These characteristics make it difficult to estimate the contact force of the Kresling cell using conventional models that depend solely on the current measurements. The same deformation measurements can correspond to different contact forces depending on the loading history. Consequently, such models cannot distinguish between loading and unloading or capture the first-cycle effect and other coupled material and geometric effects, limiting their suitability for reliable force sensing \cite{jin2022origami}.

Data-driven neural networks are suited to this task because they can learn complex coupled behaviors that are difficult to model explicitly. By taking both the deformation history and the current angle as inputs, a neural network with activation layer can capture the nonlinear relationship between these inputs and the contact force from large dataset.

Based on this idea, we designed  history-aware GRU networks that incorporate angular histories over short, medium, and long timescales, sampled at different rates as shown in Fig.~\ref{fig:archi}. To assess the importance of angular history for contact force estimation, we compared an MLP using only the current angle with GRU networks incorporating historical angle sequences. For the GRU models, we further compared a model using only short-term history with one incorporating all three timescales.

GRU network maintains a hidden state that summarizes previous inputs, while update and reset gates control how historical information is retained and incorporated into new predictions \cite{cho2014gru}. These mechanisms enable GRU networks to learn dependencies across different timescales. In our method, the input of the GRU network at each timestamp is a pair consisting of the angular displacement and a validity indicator. 

The training dataset comprises more than five hours of measurements acquired at a sampling rate of 50 Hz. The data covered a range of compression protocols, including different compression amplitudes and speeds, compression interrupted by a hold at an intermediate displacement and subsequently resumed, compression following idle periods of varying duration and so on. The independent test dataset comprises approximately 20 minutes of measurements acquired at 50 Hz. The test protocols differed from those used for training and included compression amplitudes and speeds not encountered in the training dataset.

Fig.~\ref{fig:0pressure} compares force estimation using the current angle MLP, short-history GRU, and multiscale GRU. Although all three models reproduce the overall cyclic force variation (Fig.~\ref{fig:0pressure}b), incorporating angle history improves estimation accuracy substantially. On the independent test set, the current angle model achieved an MAE of 1.28~N and an RMSE of 1.56 N. These errors decreased to 0.83~N and 0.99~N with short history input, and further to 0.59~N and 0.76~N with multiscale history (Fig.~\ref{fig:0pressure}c). The multiscale model therefore reduced test RMSE by approximately 52\% relative to the current angle model and 23\% relative to the short history model. The first-cycle detail in Fig.~\ref{fig:0pressure}d shows that the multiscale GRU captures differences between loading and unloading, as well as between the first and subsequent cycles. These results support the use of multiscale angle history to capture the history-dependent mechanical response of the Kresling sensor.

\subsection{Positive Gauge Pressure: Force Sensing with Tunable Axial Stiffness}
Following the comparison at zero gauge pressure, the multiscale GRU architecture was adopted to evaluate force sensing under positive pressure. The Kresling cell was tested at gauge pressures of 0, +10, and +20 kPa to investigate how pneumatic stiffening affects its response and force estimation. For each positive pressure, we train a new neural network based on the same method as zero gauge pressure using more than 5 hours data and test it on an independent dataset. To evaluate how pneumatic pressure tunes the force response within a limited compression stroke, we compare the three pressure conditions over the same 0–3~mm interval. Because the force–compression response is nonlinear, we use the average axial loading stiffness over this interval to quantify the pressure-dependent stiffening.

\begin{figure*}[t]
    \centering
    \includegraphics[width=\textwidth]{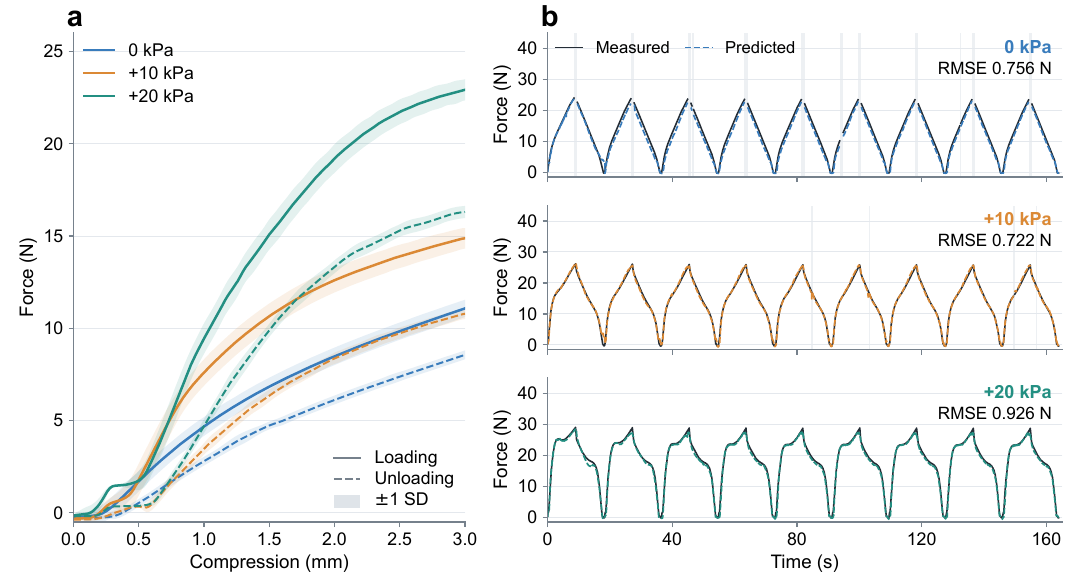}
    \caption{Tunable axial stiffness and force estimation performance under positive gauge pressure. (a) Force–compression curves at gauge pressures of 0, +10, and +20 kPa. (b) Measured and predicted force trajectories using pressure-specific multiscale GRU models.}
    \label{fig:pos}
\end{figure*}

Fig.~\ref{fig:pos}a shows the force–compression curves at the three pressure levels, with solid and dashed lines indicating loading and unloading, respectively, and shaded bands representing one standard deviation. Increasing the internal pressure increases the force required to achieve the same compression. At approximately 3 mm compression, the loading force increases from approximately 11~N at 0 kPa to 15~N at +10 kPa and 23~N at +20 kPa. During loading, the average axial loading stiffness increased from 3.78 N/mm at 0 kPa to 5.07 N/mm at +10 kPa and 7.69 N/mm at +20 kPa, corresponding to increases of 34.3\% and 103.7\% relative to zero gauge pressure.

Fig.~\ref{fig:pos}b compares the measured and predicted forces on test dataset. The predictions closely follow the measured trajectories during both loading and unloading, with test RMSEs of 0.756 N, 0.722 N, and 0.926~N at 0, +10, and +20 kPa, respectively. These results demonstrate that pneumatic pressure can modulate the axial stiffness and measurement range within a prescribed compression interval for the Kresling cell while the multiscale GRU maintains force estimation accuracy.

\subsection{Negative Gauge Pressure: Self-sensing Actuation}

\begin{figure}[t]
    \centering
    \includegraphics[width=\columnwidth]{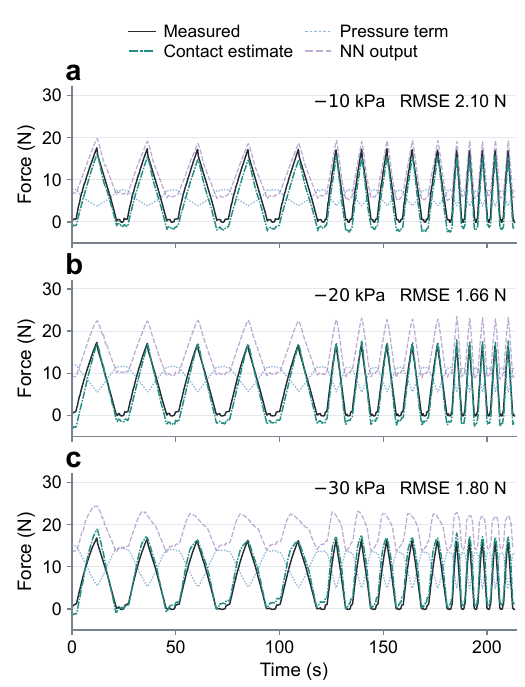}
    \caption{Contact force estimation during negative pressure actuation at nominal gauge pressures of (a) -10, (b) -20, and (c) -30 kPa. Each panel compares the measured contact force, output from the baseline model, pressure-equivalent force, and compensated contact force estimate. Compensation is applied without retraining the network.}
    \label{fig:neg}
\end{figure}

Under negative internal gauge pressure, the pressure difference drives contraction of the Kresling cell, while the magnetic encoder records the resulting rotation. In this mode, deformation arises from both pneumatic actuation and external contact force, so the neural network output cannot be interpreted directly as contact force. We therefore combined the multiscale GRU trained at zero gauge pressure with a pressure compensation model, without retraining the network on any negative pressure data.

Let $\theta$ denote the rotation relative to the initial reference angle in radians, $u$ the positive axial compression displacement, and $P=p_{\mathrm{out}}-p_{\mathrm{in}}>0$ the vacuum pressure difference. Under quasi-static conditions, the total mechanical energy absorbed by the structure is supplied by contact force work and pressure work:
\begin{equation}
\begin{aligned}
E_{\mathrm{tot}}&=W_{\mathrm{contact}}+W_p\\
&=\int F_{\mathrm{contact}}\,du-\int P\,dV,
\end{aligned}
\label{eq:energy_balance}
\end{equation}
where $V$ is the chamber volume and the integrals are evaluated along the deformation history. Here, $E_{\mathrm{tot}}$ includes both elastic energy storage and history-dependent dissipation. Without assuming that $P$ is constant, differentiating along the compression path with respect to $u$ gives
\begin{equation}
\frac{dE_{\mathrm{tot}}}{du}
=F_{\mathrm{contact}}-P\frac{dV}{du}
=F_{\mathrm{contact}}-P\frac{dV/d\theta}{du/d\theta}.
\label{eq:energy_force_balance}
\end{equation}

At zero gauge pressure, the contact force supplies the mechanical work absorbed by the structure. We therefore use the baseline model output $\widehat{F}_0(t)$ to approximate $dE_{\mathrm{tot}}/du$, assuming comparable structural resistance for a given angular deformation history and state. The contact force estimate under negative pressure becomes
\begin{equation}
\widehat{F}_{\mathrm{contact}}(t)
=\widehat{F}_0(t)+P\frac{dV/d\theta}{du/d\theta}.
\label{eq:negative_contact_force}
\end{equation}
Because chamber volume decreases during compression, the pressure contribution $F_p=-P(dV/d\theta)/(du/d\theta)$ is positive and is subtracted from the network output.

The relationships $u(\theta)$ and $V(\theta)$ were initially calculated using a theoretical geometric model that assumed a constant length of $A'B$ \cite{masana2024kresling, yang2023volume}. However, the resulting discrepancies were too large for pressure compensation. Both relationships were therefore obtained from Abaqus finite element simulations and represented by fifth-order polynomial fits. Their analytical derivatives were used to calculate $du/d\theta$ and $dV/d\theta$. For the volume model, the normalized volume--angle relationship from simulation was scaled using the initial chamber volume, $V_0=7.089$~mL, calibrated by water displacement method. Further details of the simulations are provided in the Finite Element Analysis section.

Fig.~\ref{fig:neg} compares the measured contact force, baseline model output, pressure-equivalent force, and compensated contact force estimate at nominal internal gauge pressures of $-10$, $-20$, and $-30$~kPa. Each recording contained 15 compression cycles, with 2 different loading rates. After pressure compensation, the estimated contact forces followed the measured cyclic responses, with RMSEs of 2.10, 1.66, and 1.80~N, respectively. Although estimation accuracy is lower in this mode, the method still provides approximate contact force feedback in real time, which may support manipulation tasks using soft actuators.

\section{Finite Element Analysis}

The finite element analysis of the PEBA 90A Kresling cell was performed using Abaqus 2026. The material was idealized as a homogeneous, isotropic, incompressible hyperelastic solid, and its nonlinear response under large deformation was described using a three-term Ogden model. The constitutive parameters were adopted from Su et al.~\cite{su2026peba}. In SI units, the parameters were $\mu_1=-6.58160\times10^{7}\,\mathrm{Pa}$, $\mu_2=2.67339\times10^{7}\,\mathrm{Pa}$, and $\mu_3=4.50849\times10^{7}\,\mathrm{Pa}$, with corresponding dimensionless exponents $\alpha_1=1.8026$, $\alpha_2=2.1341$, and $\alpha_3=1.3961$. The material density was set to $1008\,\mathrm{kg/m^{3}}$. Full incompressibility was assumed by setting $D_1=D_2=D_3=0$. Ten-node quadratic hybrid tetrahedral elements (C3D10H) were employed to accommodate this incompressibility constraint.

Finite element analysis was performed to characterize the coupled compression and rotation of the Kresling cell and obtain the geometric relationships required for pressure compensation. The axial compression $u$, relative rotation $\theta$, and enclosed chamber volume $V$ were extracted throughout the simulated deformation. Fifth-order polynomial fits were used to represent $u(\theta)$ and the normalized volume--angle relationship. The volume model was subsequently scaled using the initial chamber volume calibrated by water displacement method.

The simulated configurations were also inspected for internal self-contact between opposing walls and folded facets. No internal self-contact was observed within the compression range retained for sensing. This check was necessary because contact between internal surfaces introduces additional constraints and force contributions that are not represented by the assumed relationship. The operating range was therefore restricted to configurations preceding internal contact, and data exceeding the selected compression limit were excluded.

\begin{figure}[t]
    \centering
    \includegraphics[width=\columnwidth]{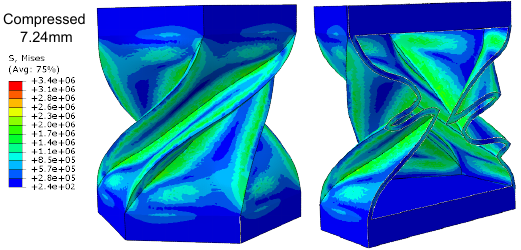}
    \caption{Finite element analysis of the von Mises stress distribution in the PEBA Kresling cell at an axial compression of 7.24 mm. The left panel shows the exterior view, and the right panel shows a cross-sectional view. Elevated stresses are concentrated around the folding regions.}
    \label{fig:fem}
\end{figure}

Stress analysis yielded a global maximum von Mises stress of 4.665~MPa over the simulated compression path. Fig.~\ref{fig:fem} illustrates the stress distribution at a representative compression of 7.24~mm, with elevated stresses concentrated around the folding regions. For comparison, Su et al. reported tensile fracture stresses for FDM printed PEBA 90A, with the lowest value occurring in the z-orientation at approximately 10.5~MPa \cite{su2026peba}. The simulated peak stress is approximately 44\% of this value, providing a preliminary indication that the stress demand remains below the published tensile fracture benchmark. This strength comparison is indicative because the simulated von Mises stress and experimentally measured engineering fracture stress are different measures. Long-term durability still requires cyclic testing to assess fatigue and damage accumulation at the folds.

\section{Conclusion \& Discussion}

This work demonstrates a 3D-printed PEBA Kresling cell that combines magnetic angular sensing, multiscale history learning, and pneumatic modulation to support three operating modes. At zero gauge pressure, the multiscale GRU estimates contact force while accounting for hysteresis and first-cycle effects. Under positive pressure, pressure-specific models maintain force estimation accuracy as pneumatic stiffening adjusts the average axial stiffness and accessible force range within a prescribed compression interval. Under negative pressure, an energy-based correction extends the baseline model to self-sensing actuation without retraining. Together, these results demonstrate the potential to integrate force feedback, adjustable compliance, and pneumatic actuation within a single origami cell.

Several limitations remain. The negative-pressure compensation method assumes quasi-static conditions, restricting its current applicability to low deformation rates. Uncertainty in the volume--angle relationship and deviations between nominal and actual chamber pressure further affect compensation accuracy. Future improvements could incorporate measured chamber pressure, experimentally calibrated geometric relationships, and dynamic effects into the estimator. More broadly, generalization to unseen loading histories and different specimens remains to be established through additional testing, together with evaluation of long-term cyclic durability.

Future work will investigate how these capabilities can support soft robotic systems under bending and combined loading. Integrating force feedback with tunable stiffness could enable soft robotic systems to adapt their mechanical response to different stages of an interaction and different payloads. Other applications include compliant grippers that adjust their force response for objects of different fragility and tactile interfaces with selectable force ranges.

We hope that this work will inspire further development of origami-based soft robots that combine adaptable mechanical properties with integrated sensing for interaction with their surroundings.




\section*{Appendix}

\subsection{Photograph of the Experimental Setup}

\begin{figure}[h]
    \centering
    \includegraphics[width=0.5\columnwidth]{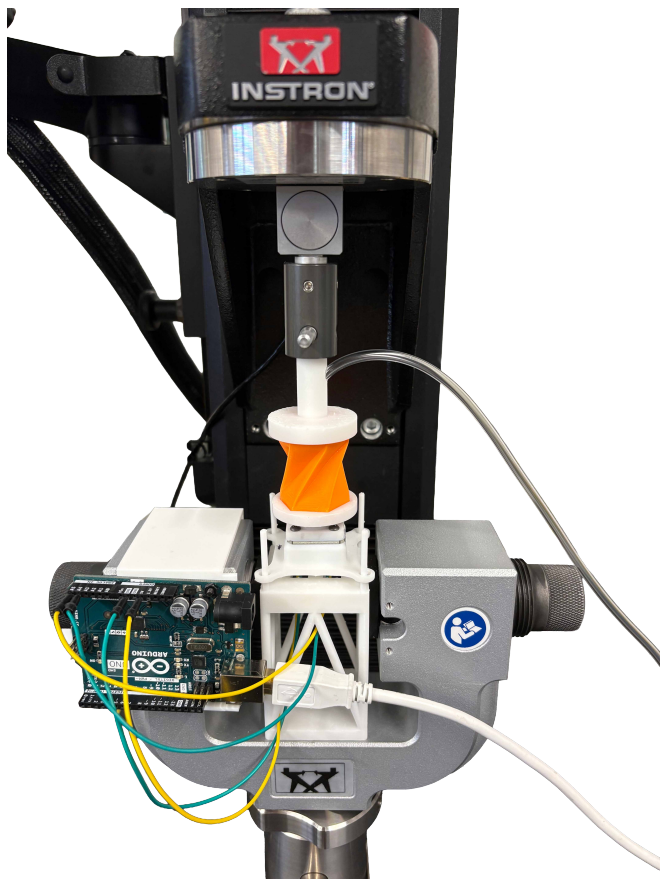}
    \caption{Photograph of the experimental platform, showing the Kresling cell mounted in the Instron testing machine with magnetic angular sensing, pneumatic tubing, and data acquisition electronics.}
    \label{fig:exp}
\end{figure}

\subsection{Polynomial Fits to Finite Element Results}
\begin{equation}
    \label{eq:displacement_angle}
    \begin{aligned}
    u(\theta)=10^{-3}\bigl(&6.937884996\theta
    -3.354149208\theta^{2}\\
    &+11.813757709\theta^{3}
    -15.184050672\theta^{4}\\
    &+7.968567917\theta^{5}\bigr)\;\mathrm{m},
    \end{aligned}
\end{equation}

\begin{equation}
    \label{eq:volume_angle}
    \begin{aligned}
    V(\theta)=V_0\bigl(&1
    -0.973237823\theta
    +1.410395874\theta^{2}\\
    &-2.429661324\theta^{3}
    +2.338355064\theta^{4}\\
    &-0.823522904\theta^{5}\bigr),
    \end{aligned}
\end{equation}
where $\theta$ is the rotation relative to the initial state in radians,
$u=h_0-h$ is the downward positive compression displacement, and
$V_0=7.089\times10^{-6}\,\mathrm{m}^{3}$ by water displacement method.
Both relations apply over $0\leq\theta\leq1.1865304\,\mathrm{rad}$.

\section*{Acknowledgment}
The authors acknowledge the assistance of OpenAI Codex in developing the code framework and scripts for angular data acquisition, model training, validation, and testing, as well as in conducting literature searches.



\begin{thebibliography}{99}

\bibitem{zhao2016optoelectronic}
H. Zhao, K. O'Brien, S. Li, and R. F. Shepherd, ``Optoelectronically innervated soft prosthetic hand via stretchable optical waveguides,'' \textit{Science Robotics}, vol. 1, no. 1, Art. no. eaai7529, 2016, doi: 10.1126/scirobotics.aai7529.

\bibitem{truby2018somatosensitive}
R. L. Truby, M. Wehner, A. K. Grosskopf, D. M. Vogt, S. G. M. Uzel, R. J. Wood, and J. A. Lewis, ``Soft somatosensitive actuators via embedded 3D printing,'' \textit{Advanced Materials}, vol. 30, Art. no. 1706383, 2018, doi: 10.1002/adma.201706383.

\bibitem{shih2019sensors}
B. Shih, C. Christianson, K. Gillespie, S. Lee, J. Mayeda, Z. Huo, and M. T. Tolley, ``Design considerations for 3D printed, soft, multimaterial resistive sensors for soft robotics,'' \textit{Frontiers in Robotics and AI}, vol. 6, Art. no. 30, 2019, doi: 10.3389/frobt.2019.00030.

\bibitem{jin2022origami}
T. Jin, L. Li, T. Wang, G. Wang, J. Cai, Y. Tian, and Q. Zhang, ``Origami-inspired soft actuators for stimulus perception and crawling robot applications,'' \textit{IEEE Transactions on Robotics}, vol. 38, no. 2, pp. 748--764, 2022, doi: 10.1109/TRO.2021.3096644.

\bibitem{masana2024kresling}
R. Masana, A. S. Dalaq, S. Khazaaleh, and M. F. Daqaq, ``The Kresling origami spring: A review and assessment,'' \textit{Smart Materials and Structures}, vol. 33, no. 4, Art. no. 043002, 2024, doi: 10.1088/1361-665X/ad2f6f.

\bibitem{hanson2024fold}
N. Hanson, I. A. Mensah, S. F. Roberts, J. Healey, C. Wu, and K. L. Dorsey, ``Controlling the fold: proprioceptive feedback in a soft origami robot,'' \textit{Frontiers in Robotics and AI}, vol. 11, Art. no. 1396082, 2024, doi: 10.3389/frobt.2024.1396082.

\bibitem{oliveira2026orisense}
H. de Souza Oliveira, X. Li, M. Jafarpour, and E. Milana, ``Ori-Sense: origami capacitive sensing for soft robotic applications,'' arXiv:2602.18379, 2026, doi: 10.48550/arXiv.2602.18379.

\bibitem{fan2024vacuum}
W. Fan, J. Wang, Z. Zhang, G. Chen, and H. Wang, ``Vacuum-driven parallel continuum robots with self-sensing origami linkages,'' \textit{IEEE/ASME Transactions on Mechatronics}, vol. 29, no. 5, pp. 3370--3380, 2024, doi: 10.1109/TMECH.2023.3340152.

\bibitem{fan2025sparc}
W. Fan, J. Wang, H. Zhang, Q. Wu, Q. Hu, G. Chen, H. Wang, and X. Huang, ``SPARC: A soft, proprioceptive, agile robot for 3D climbing and exploration with precise trajectory following,'' \textit{Advanced Science}, vol. 12, no. 41, Art. no. e10382, 2025, doi: 10.1002/advs.202510382.

\bibitem{wei2025biodegradable}
P. Wei et al., ``Biodegradable origami enables closed-loop sustainable robotic systems,'' \textit{Science Advances}, vol. 11, no. 6, Art. no. eads0217, 2025, doi: 10.1126/sciadv.ads0217.

\bibitem{wang2023coiling}
Y.-H. Wang, Z.-L. Xu, Y. Wang, R. Huan, H. Jiang, and K.-C. Chuang, ``Coiling an optical fiber for long-range dynamic displacement and force sensing,'' \textit{Extreme Mechanics Letters}, vol. 62, Art. no. 102032, 2023, doi: 10.1016/j.eml.2023.102032.

\bibitem{kim2026bifurcation}
Y.-S. Kim, J. Yoon, S. Cho, J. Kim, and D. Choi, ``Bifurcation-gated Kresling origami triboelectric mechanoreceptor enabling self-powered multimodal tactile sensing for intelligent human-robot interaction,'' \textit{Nano Energy}, vol. 157, Art. no. 112280, 2026, doi: 10.1016/j.nanoen.2026.112280.

\bibitem{huang2022modular}
J. Huang, J. Zhou, Z. Wang, J. Law, H. Cao, Y. Li, H. Wang, and Y. Liu, ``Modular origami soft robot with the perception of interaction force and body configuration,'' \textit{Advanced Intelligent Systems}, vol. 4, no. 9, Art. no. 2200081, 2022, doi: 10.1002/aisy.202200081.

\bibitem{buso2020contact}
A. Buso, R. B. N. Scharff, E. L. Doubrovski, J. Wu, C. C. L. Wang, and P. Vink, ``Soft robotic module for sensing and controlling contact force,'' in \textit{Proc. IEEE International Conference on Soft Robotics (RoboSoft)}, 2020, pp. 70--75, doi: 10.1109/RoboSoft48309.2020.9115970.

\bibitem{mora2025landscape}
S. Mora, N. M. Pugno, and D. Misseroni, ``Programming the energy landscape of 3D-printed Kresling origami via crease geometry and viscosity,'' \textit{Extreme Mechanics Letters}, vol. 77, Art. no. 102314, 2025, doi: 10.1016/j.eml.2025.102314.

\bibitem{han2018microfluidic}
S. Han, T. Kim, D. Kim, Y.-L. Park, and S. Jo, ``Use of deep learning for characterization of microfluidic soft sensors,'' \textit{IEEE Robotics and Automation Letters}, vol. 3, no. 2, pp. 873--880, 2018, doi: 10.1109/LRA.2018.2792684.

\bibitem{thuruthel2019perception}
T. G. Thuruthel, B. Shih, C. Laschi, and M. T. Tolley, ``Soft robot perception using embedded soft sensors and recurrent neural networks,'' \textit{Science Robotics}, vol. 4, no. 26, Art. no. eaav1488, 2019, doi: 10.1126/scirobotics.aav1488.

\bibitem{truby2020distributed}
R. L. Truby, C. Della Santina, and D. Rus, ``Distributed proprioception of 3D configuration in soft, sensorized robots via deep learning,'' \textit{IEEE Robotics and Automation Letters}, vol. 5, no. 2, pp. 3299--3306, 2020, doi: 10.1109/LRA.2020.2976320.

\bibitem{cho2014gru}
K. Cho, B. van Merri{\"e}nboer, C. Gulcehre, D. Bahdanau, F. Bougares, H. Schwenk, and Y. Bengio, ``Learning phrase representations using RNN encoder--decoder for statistical machine translation,'' in \textit{Proc. Conf. Empirical Methods in Natural Language Processing (EMNLP)}, 2014, pp. 1724--1734, doi: 10.3115/v1/D14-1179.

\bibitem{su2026peba}
R. Su, J. Du, C. Jia, Z. Gan, and Z. Zhang, ``3D printing elastic metamaterials and actuators with high resilience via polyether block amide (PEBA),'' \textit{Extreme Mechanics Letters}, vol. 84, Art. no. 102465, 2026, doi: 10.1016/j.eml.2026.102465.

\bibitem{mullins1969}
L. Mullins, ``Softening of rubber by deformation,'' \textit{Rubber Chemistry and Technology}, vol. 42, no. 1, pp. 339--362, 1969, doi: 10.5254/1.3539210.

\bibitem{yang2023volume}
M. Yang, J. Defillion, F. Scarpa, and M. Schenk, ``Volume optimisation of multi-stable origami bellows for deployable space habitats,'' \textit{Acta Mechanica Solida Sinica}, vol. 36, no. 4, pp. 514--530, 2023, doi: 10.1007/s10338-023-00401-3.

\end{thebibliography}
\end{document}